# The Phylogenetic Indian Buffet Process:
# A Non-Exchangeable Nonparametric Prior for Latent Features

**Kurt T. Miller**
EECS
University of California
Berkeley, CA 94720
tadayuki@cs.berkeley.edu

**Thomas L. Griffiths**
Psychology and Cognitive Science
University of California
Berkeley, CA 94720
tom_griffiths@berkeley.edu

**Michael I. Jordan**
EECS and Statistics
University of California
Berkeley, CA 94720
jordan@cs.berkeley.edu

## Abstract

Nonparametric Bayesian models are often based on the assumption that the objects being modeled are exchangeable. While appropriate in some applications (e.g., bag-of-words models for documents), exchangeability is sometimes assumed simply for computational reasons; non-exchangeable models might be a better choice for applications based on subject matter. Drawing on ideas from graphical models and phylogenetics, we describe a non-exchangeable prior for a class of nonparametric latent feature models that is nearly as efficient computationally as its exchangeable counterpart. Our model is applicable to the general setting in which the dependencies between objects can be expressed using a tree, where edge lengths indicate the strength of relationships. We demonstrate an application to modeling probabilistic choice.

## 1 INTRODUCTION

Nonparametric Bayesian analysis provides a way to define probabilistic models in which structural aspects of the model, such as the number of classes or features possessed by a set of objects, are treated as unknown, unbounded and random, and thus viewed as part and parcel of the posterior inference problem. This elegant treatment of structural uncertainty has led to increasing interest in nonparametric Bayesian approaches as alternatives to model selection procedures.

Nonparametric Bayesian methods are based on prior distributions that are defined on infinite collections of random variables—i.e., prior distributions expressed as general stochastic processes. This generality provides a rich language in which to express prior knowledge. In practice, however, the need to integrate over these stochastic processes at inference time imposes a strong constraint on the kinds of models that can be considered. In particular, nonparametric Bayesian models are often based on an assumption of *exchangeability*—the joint probability of the set of entities being modeled by the prior is assumed to be invariant to permutation. A particular example of exchangeability is the "bag-of-words" assumption widely used in the modeling of document collections.

In this paper we aim to extend the range of nonparametric Bayesian modeling by presenting a non-exchangeable prior for a class of nonparametric latent feature models. Our point of departure is the *Indian buffet process* (IBP), a generative process that defines a prior on sparse binary matrices (Griffiths and Ghahramani, 2006). This process, which will be described in more depth later, can be understood through a culinary metaphor in which diners sequentially enter a buffet line and select which dishes to try. As each person moves through the buffet line, they try each previously sampled dish with probability proportional to the number of people who have already tried it. This process can be shown to be exchangeable from the fact that it is obtained as a conditionally-independent set of draws from a Bernoulli process with parameters drawn from an underlying stochastic process known as the beta process (Thibaux and Jordan, 2007).

To obtain a useful non-exchangeable, nonparametric model while retaining the computational tractability of the IBP, we consider a model in which relationships among the diners are expressed by a tree. In this stochastic process—the *Phylogenetic Indian Buffet Process* (pIBP)—the probability that a diner chooses a dish in the buffet line depends not only on the number of previous diners who have chosen that dish, but also on how closely related the current diner is to each of the previous diners. We exploit efficient probabilistic calculations on trees (Pearl, 1988) to obtain a tractable algorithm for taking relatedness into account when computing posterior updates under the pIBP prior.

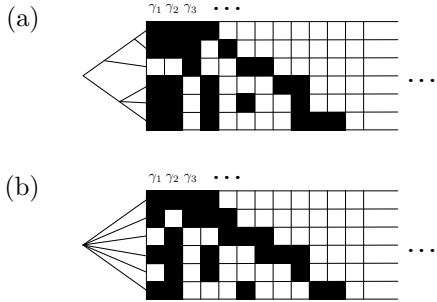

Figure 1: The Phylogenetic Indian Buffet Process. (a) A tree expresses dependencies among featural representations of objects. (b) The Indian Buffet Process is a special case of the pIBP where all branches meet at the root.

The tree representation is a rich language for expressing non-exchangeability. In particular, factorial and nested models of the kind used in experimental design are readily expressed as trees. Group structure as used in the hierarchical Dirichlet process (Teh et al., 2005) and hierarchical beta process (Thibaux and Jordan, 2007) can also be expressed as trees, as can a variety of other partially exchangeable models (Diaconis, 1988). In biological data analysis we may be able to exploit known phylogenetic or genealogical relationships among species or characters. More generally we may have similarity data available for a set of objects that can be used to build a tree representation. The pIBP uses such representations to induce a prior on featural representations such that objects that are related in the tree will tend to share features (see Figure 1).

It is important to distinguish our approach from previous nonparametric Bayesian work based on random trees (Neal, 2001; Teh et al., 2008). In that work, trees are averaged over and objects remain exchangeable. We are conditioning on a known, fixed tree and our objects are not exchangeable.

While we develop the pIBP in the context of the IBP for concreteness, the idea is actually much broader. The key is that the use of a tree to express relationships among non-exchangeable random variables allows us to exploit the sum-product algorithm in defining the updates for an MCMC sampler. This insight extends the scope of nonparametric Bayesian models without significantly increasing the computational burden associated with inference.

The paper is organized as follows. Section 2 presents a short review of the IBP and then provides a detailed description of the pIBP. Section 3 discusses MCMC inference in models using the pIBP as a prior. Section 4 presents an application of the pIBP to models of human choice, and shows how combining nonparametric methods with a tree-based prior improves performance. Section 5 presents our conclusions.

## 2 THE PHYLOGENETIC INDIAN BUFFET PROCESS

Griffiths and Ghahramani (2006) proposed the Indian Buffet Process as a prior distribution on sparse binary matrices $Z$, where the rows of $Z$ correspond to objects and the columns of $Z$ correspond to a set of features or attributes describing these objects. As with other nonparametric Bayesian models, the IBP can be interpreted through a culinary metaphor. In this metaphor, objects correspond to people and features correspond to an infinite array of dishes at an Indian buffet. The first person samples Poisson($\alpha$) dishes, where $\alpha$ is a parameter. The $i^{\text{th}}$ customer then samples all previously sampled dishes in proportion to the number of people who have already sampled those dishes, and Poisson($\alpha/i$) new dishes. This process defines an exchangeable distribution on equivalence classes of $Z$, the binary matrix with a one at cell $(i, k)$ when the $i^{\text{th}}$ customer chooses the $k^{\text{th}}$ dish. Figure 1(b) shows a matrix generated from this process.

The IBP can be derived as the infinite limit of a beta-Bernoulli model. Consider a finite model in which there are $K$ features, and let the probability that an object possesses feature $k$ be Bernoulli($\pi_k$). Endowing $\pi_k$ with a Beta($\alpha/K, 1$) distribution, we obtain the IBP in the limit as $K \to \infty$. The Phylogenetic Indian Buffet Process uses a similar construction. As with the IBP, we will use the term "pIBP" to refer to both the distribution on binary matrices as well as a generative process which induces this distribution. We first describe the distribution on finite matrices at the heart of the pIBP, and then describe the process that results by letting the number of features go to infinity.

### 2.1 A PRIOR ON FINITE MATRICES

We begin by defining a generative process for $Z$, an $N \times K$ binary matrix, where $N$ is the number of objects and $K$ is a fixed, finite number of features. We use the following notation. Let $z_{ik}$ denote the $(i, k)$ entry of $Z$, let $z_k$ be the $k^{\text{th}}$ column of $Z$, let $z_{(-i)k}$ denote the entries of $z_k$ except $z_{ik}$, and let $Z_{-(ik)}$ denote the entries of the full $Z$ matrix except $z_{ik}$.

As in the IBP, we associate a parameter $\pi_k$ to each column, chosen from a Beta($\alpha/K, 1$) prior distribution, where $\alpha$ is a hyperparameter. Given $\pi_k$, the marginal probability that any particular entry in column $k$ is one is equal to $\pi_k$. Columns are generated independently. We diverge from the IBP, however, in the specification of the joint probability distribution for the column $z_k$. In the IBP, the entries of $z_k$ are chosen independently given $\pi_k$. In the pIBP, the entries of $z_k$ are dependent, with the pattern of dependence captured by a stochastic process on a rooted tree similar to models used in phylogenetics. In this tree, the $N$ objects being modeled are at the leaves, and lengths are assigned to edges in such a way that the total edge length from the root to any leaf is equal to one. To generate the entries of the $k^{\text{th}}$ column, we proceed as follows. Assign the value zero to the root node of the tree. Along any path from the root to a leaf, let this value change to a one along any edge of length $t$ with exponential rate $\gamma_k t$ where $\gamma_k = -\log(1 - \pi_k)$. That is, along an edge of length $t$, let the probability of changing from a zero to a one be $1 - \exp(-\gamma_k t)$. Once the value has changed to a one along any path from the root, all leaves below that point are assigned the value one.

The parameterization $\gamma_k = -\log(1 - \pi_k)$ is convenient because it ensures that $\pi_k$ remains the marginal probability that any single feature is equal to one. To see this, simply note that since every leaf node is at distance one from the root, for any entry in the matrix,

$$p(z_{ik} = 1|\pi_k) = 1 - \exp(-(-\log(1 - \pi_k))) = \pi_k$$

which also guarantees that we recover the beta-Bernoulli model in the special case where all branches join at the root, as in Figure 1 (b). It is a simple corollary that any set of objects characterized by a set of branches that meet at a single point will be exchangeable within that set, meaning that the tree can be used to capture notions of partial exchangeability.

## 2.2 CONDITIONAL PROBABILITIES

Now that we have defined the generative model on finite matrices, we show how to evaluate conditional probabilities in this model. We treat the tree as a directed graph with variables at each of the interior nodes and $z_{ik}$ at each leaf $i$. Then, given $\pi_k$, or equivalently $\gamma_k$, if there is a length $t$ edge from a parent node $x$ to a child node $y$, we have

$$\begin{aligned} p(y=0|x=0,\gamma_k) &= \exp(-\gamma_k t) \\ p(y=1|x=0,\gamma_k) &= 1 - \exp(-\gamma_k t) \\ p(y=0|x=1,\gamma_k) &= 0 \\ p(y=1|x=1,\gamma_k) &= 1 \end{aligned}$$

as the conditional probabilities that define a tree-structured graphical model.

Expressing this process as a graphical model makes it possible to efficiently compute various conditional probabilities that are relevant for posterior inference. Specifically, we will need to evaluate

$$p(z_{ik}|z_{(-i)k}, \pi_k) \quad (1)$$

for $z_{ik} \in \{0, 1\}$ and

$$p(z_{(-i)k}|z_{ik}, \pi_k), \quad (2)$$

which are trivial in the beta-Bernoulli model due to the conditional independence of $z_{ik}$, but more challenging in the pIBP where $z_{ik}$ are no longer conditionally independent. To compute (1), we use the sum-product algorithm (Pearl, 1988). In order to calculate (2), we use the chain rule of probability to get a set of terms similar to (1), the difference being that the posterior in each term is conditional only on a subset of the other variables. Each term can be reduced to a simple sum-product calculation by marginalizing over all variables that do not appear in that term, which can be done easily since all variables appear at the leaves of the tree. Both (1) and (2) can be calculated in $O(N)$ time by a dynamic program.

## 2.3 A GENERATIVE PROCESS

The pIBP can be described as a sequential generative process that arises when we let $K$ go to infinity in the distribution derived in Section 2.1. This process can again be understood in terms of a culinary metaphor, in which each row of $Z$ is viewed as the choices made by a diner in a buffet line, and in which we specify how each diner chooses their dishes based on the dishes chosen by previous diners. We overview this process here, leaving detailed mathematical derivations for later sections.

Consider a large extended family that is about to choose dishes at a buffet. Assume that we are given a tree describing the genealogical relationships of the family members and assume that dining preferences are related to genealogy. In particular, family members who are more closely related have more similar preferences. Therefore, as each diner moves through the buffet line, their choice of dishes will be more dependent on the selections of previous diners who are closely related to them and less dependent on the selections of other diners.

The pIBP generative process is specified as follows. The first diner (arbitrarily chosen) starts at the head of a buffet line that has infinitely many dishes. This person tries Poisson($\alpha$) dishes and also adds a brief annotation to each of these dishes, $\pi_k$, drawn uniformly from $[0,1]$. This note, through its previously described equivalent representation, $\gamma_k = -\log(1 - \pi_k)$, will allow us to efficiently compute the probability that subsequent diners choose the $k^{\text{th}}$ dish using the sum-product algorithm.

Each subsequent diner enters the buffet line and based on the annotations attached to the dishes as well as the identity of previous diners, samples the $k^{\text{th}}$ dish according to the probability (1) where $z_{(-i)k}$ indicates which of the previous diners have chosen the $k^{\text{th}}$ dish. Through the stochastic process on the tree, if closely related diners have tried a dish, the current diner is more likely to also sample it. The preferences of all diners who have not entered the buffet line are ignored, which can be done by only performing sum-product on the minimal subtree from the root containing the current diner and all previous diners.

Each diner also samples a number of new dishes. If $t_i$ is the length of the branch connecting diner $i$ to the rest of the minimal subtree just described and $\sum t$ is the total length of the rest of this subtree, then diner $i$ tries Poisson($\alpha(\psi(\sum t + t_i + 1) - \psi(\sum t + 1))$) new dishes, where $\psi(\cdot)$ is the digamma function. They also add an annotation, $\pi_k$, to each of the new dishes that will be used for future inferences, where the density of $\pi_k$ is proportional to $\left(1 - (1-\pi_k)^{t_i}\right)(1-\pi_k)^{\sum t} \pi_k^{-1}$.

This process repeats until all diners have gone through the buffet line, defining a matrix $Z$ as in the IBP. Though this process is not exchangeable, we can let any family member go first and get the same marginal distribution. This means that each row of $Z$ has a Poisson($\alpha$) number of non-zero columns, yielding a sparse matrix as in the IBP. The IBP is the special case of the pIBP corresponding to the tree shown in Figure 1 (b); this fact can be proved by using identities of the digamma function on the integers.

## 3 INFERENCE BY MCMC

We now consider how to perform posterior inference in models using the pIBP as a prior. As with the IBP, exact inference is intractable, but the model is amenable to approximate inference via Markov chain Monte Carlo (MCMC) (Robert and Casella, 2004). An important point is that even though we are dealing with a potentially infinite number of columns in $Z$, we only need to keep track of the non-zero columns, a property shared by other nonparametric Bayesian models. Henceforth, let $Z$ refer to all the non-zero columns of the matrix. The fact that the number of ones in any given row has a Poisson distribution means that the number of non-zero columns is finite (with probability one) and generally small.

Given the matrix $Z$, we assume that data $X$ are generated through a likelihood function $p(X|Z)$. The likelihood may introduce additional parameters that must be sampled as part of the overall MCMC procedure; we will not discuss such parameters in our presentation.

Unlike the IBP, where $\pi_k$ can always be integrated out, inference in the pIBP requires treating $\pi_k$ as an auxiliary variable, sampling it when needed and integrating it out when possible. By sampling $\pi_k$ for non-zero columns of $Z$ as opposed to integrating it out, we are able to exploit the sum-product algorithm as described in Section 2.2. Updating $\pi_k$ and all $z_{ik}$ for each column takes $O(N + mN)$ time where $m$ is the total number of times a $z_{ik}$ in column $k$ changes value. Once the chain has mixed well $m$ is typically small, so time complexity is only slightly worse than that of the IBP, which is $O(N)$.

Given an initial choice of the non-zero columns of $Z$ and the corresponding $\pi_k$ for each of these columns, we construct a Markov chain where at each step, we only need to sample each variable from its conditional distribution given all others. We now describe how to sample each of the variables, first considering the variables for "old" columns—those with non-zero entries—and then turning to the addition of "new" columns.

### 3.1 SAMPLING $z_{ik}$ FOR OLD COLUMNS

The probability of each $z_{ik}$ given all other variables is

$$p(z_{ik}|Z_{-(ik)}, \pi_k, X, \alpha) \propto p(X|Z_{-(ik)}, z_{ik})p(z_{ik}|z_{(-i)k}, \pi_k), \quad (3)$$

where the first term is the probability of $X$ given a full assignment of the parameters and depends on the specific model being used. The term $p(z_{ik}|z_{(-i)k}, \pi_k)$ can be computed efficiently using the sum-product algorithm as described in Section 2.2. By appropriately caching messages from the sum-product algorithm, this evaluation can be reduced to $O(1)$ time. We evaluate (3) for $z_{ik} = 0$ and $z_{ik} = 1$ and sample $z_{ik}$ from the corresponding posterior distribution. If the value of $z_{ik}$ changes, we then update the messages for sum-product in $O(N)$ time. If a column ever becomes entirely zero, we drop it from $Z$.

### 3.2 SAMPLING $\pi_k$ FOR OLD COLUMNS

We only sample $\pi_k$ for the old columns of $Z$, a fact that will be useful in subsequent calculations. The posterior distribution of each $\pi_k$ is independent of all other $\pi_k$ and depends only on the $k^{\text{th}}$ column of $Z$.

When resampling $\pi_k$, let $z_{ik}$ be a non-zero entry in the $k^{\text{th}}$ column. Then,

$$p(\pi_k|z_k, \alpha) \propto p(z_{(-i)k}|\pi_k, z_{ik})p(\pi_k|z_{ik}, \alpha). \quad (4)$$

Section 2.2 describes how to compute $p(z_{(-i)k}|\pi_k, z_{ik})$ efficiently in $O(N)$ time using the sum-product algo-

rithm. In order to obtain $p(\pi_k|z_{ik}, \alpha)$, we compute

$$p(\pi_k|z_{ik}, \alpha) \propto \underbrace{p(z_{ik}|\pi_k)}_{\sim \text{Bernoulli}(\pi_k)} \underbrace{p(\pi_k|\alpha)}_{\sim \text{Beta}(\frac{\alpha}{K},1)}$$

$$\propto \text{Beta}\left(1 + \frac{\alpha}{K}, 1\right)$$

$$\stackrel{K\to\infty}{\to} \text{Beta}(1,1) = \mathbb{1}_{\{\pi_k \in [0,1]\}}.$$

We see that we can evaluate $p(\pi_k|z_k, \alpha)$ up to a normalizing constant efficiently, so we can sample the new $\pi_k$ using a Metropolis-Hastings step.

Specifically, given a proposed value of $\pi'$ for $\pi_k$, the ratio of the posterior probabilities of $\pi'$ and $\pi_k$ is

$$\frac{p(\pi'|z_k, \alpha)}{p(\pi_k|z_k, \alpha)} = \frac{p(z_{(-i)k}|\pi', z_{ik})p(\pi'|z_{ik}, \alpha)}{p(z_{(-i)k}|\pi_k, z_{ik})p(\pi_k|z_{ik}, \alpha)}$$

$$= \frac{p(z_{(-i)k}|\pi', z_{ik})}{p(z_{(-i)k}|\pi_k, z_{ik})} \mathbb{1}_{\{\pi' \in [0,1]\}},$$

so if we use $q(\pi'|\pi_k)$ as the proposal distribution, then the Metropolis-Hastings acceptance ratio is

$$\min\left[1, \frac{q(\pi_k|\pi')}{q(\pi'|\pi_k)} \frac{p(z_{(-i)k}|\pi', z_{ik})}{p(z_{(-i)k}|\pi_k, z_{ik})} \mathbb{1}_{\{\pi' \in [0,1]\}}\right]. \quad (5)$$

There are many options for $q$. In our experiments, we used $q(\pi'|\pi_k) \sim \mathcal{N}(\pi_k, \sigma_k^2)$ where $\sigma_k^2 = c \cdot \pi_k(1-\pi_k) + \delta$ with $c = 0.06$ and $\delta = 0.08$.

### 3.3 SAMPLING THE NEW COLUMNS

In addition to sampling the values of $z_{ik}$ in old columns, we need to consider the possibility that one of the infinitely many all-zero columns has a single entry that becomes a one. As mentioned in Section 3.2, we only sample values of $\pi$ for non-zero columns, so when sampling new columns, we must integrate out $\pi$.

For finite $K$, we can directly compute the probability $p(z_{ik} = 1|z_{(-i)k} = 0)$ for each all-zero column. If $t_i$ is the length of the edge that ends at the $i^{\text{th}}$ object and $\sum t$ is the total length of all other edges in the tree, then we get

$$p(z_{ik} = 1|z_{(-i)k} = 0)$$

$$\propto \int_0^1 p(z_{ik} = 1|z_{(-i)k} = 0, \pi) p(z_{(-i)k} = 0|\pi) p(\pi) d\pi$$

$$\propto \int_0^1 (1-(1-\pi)^{t_i})(1-\pi)^{\sum t} \pi^{\alpha/K - 1} d\pi$$

$$= \frac{\Gamma(\alpha/K)\Gamma(\sum t + 1)}{\Gamma(\sum t + \alpha/K + 1)} - \frac{\Gamma(\alpha/K)\Gamma(\sum t + t_i + 1)}{\Gamma(\sum t + t_i + \alpha/K + 1)}$$

where $\Gamma(\cdot)$ is the gamma function and similarly

$$p(z_{ik} = 0|z_{(-i)k} = 0) \propto \frac{\Gamma(\alpha/K)\Gamma(\sum t + t_i + 1)}{\Gamma(\sum t + t_i + \alpha/K + 1)}$$

which gives us

$$p(z_{ik} = 1|z_{(-i)k} = 0)$$

$$= 1 - \frac{\Gamma(\sum t + t_i + 1)}{\Gamma(\sum t + 1)} \frac{\Gamma(\sum t + \alpha/K + 1)}{\Gamma(\sum t + t_i + \alpha/K + 1)}.$$

Therefore, the event that we sample $z_{ik} = 1$ in any particular all-zero column is Bernoulli with the above probability. Treating the value $\alpha/K$ as a variable in the equation for $p(z_{ik} = 1|z_{(-i)k} = 0)$ and doing a first-order Taylor expansion about zero, we get

$$p(z_{ik} = 1|z_{(-i)k} = 0)$$

$$= \frac{\alpha}{K}\left(\psi\left(\sum t + t_i + 1\right) - \psi\left(\sum t + 1\right)\right) + o\left(\frac{1}{K}\right),$$

where $\psi(\cdot)$ is the digamma function.

As $K$ grows, the probability that any particular all-zero column becomes non-zero goes to zero. On the other hand, we have a growing number of these Bernoulli variables. Using the fact that Binomial$\left(K, \frac{\lambda}{K}\right) \stackrel{K\to\infty}{\to}$ Poisson$(\lambda)$, then we get that for each row $i$, the number of new non-zero columns with a one in the $i^{\text{th}}$ row is distributed

$$\text{Poisson}\left(\alpha\left(\psi\left(\sum t + t_i + 1\right) - \psi\left(\sum t + 1\right)\right)\right). \quad (6)$$

Putting this all together, instead of sampling $z_{ik}$ for each of the infinitely many all-zero columns individually, we sample $K_i^{\text{new}}$, the number of these columns which will have $z_{ik} = 1$. The distribution of $K_i^{\text{new}}$ is

$$p(K_i^{\text{new}}|X, Z, \alpha) \propto P(X|Z_{\text{new}})P(K_i^{\text{new}}|\alpha), \quad (7)$$

where $P(K_i^{\text{new}}|\alpha)$ is given by Equation (6). To compute $P(X|Z_{\text{new}})$, we must augment $Z$ with $K_i^{\text{new}}$ new columns that have a one in only the $i^{\text{th}}$ row; this yields $Z_{\text{new}}$. Though $K_i^{\text{new}}$ can be arbitrarily large, (7) decays rapidly as $K_i^{\text{new}}$ grows, so we can truncate our evaluation at a finite number of columns and sample $K_i^{\text{new}}$ from the corresponding multinomial.

### 3.4 SAMPLING $\pi_k$ FOR NEW COLUMNS

For each of the new columns generated in the previous step, we must sample an initial value of $\pi_k$. Using the same notation as before for edge lengths, if we are sampling $\pi_k$ for a new column in which only the $i^{\text{th}}$ element is non-zero, then we are sampling from

$$p(\pi_k|z_k) = p(\pi_k|z_{(-i)k} = 0, z_{ik} = 1)$$

$$\propto \left(1 - (1-\pi_k)^{t_i}\right)(1-\pi_k)^{\sum t} \pi_k^{-1}. (8)$$

To obtain a sample from this distribution, we use the Metropolis-Hastings sampler from Section 3.2 where we replace equation (4) with (8).

## 3.5 SAMPLING $\alpha$

Following Görür et al. (2006), we place a gamma prior, $\mathcal{G}(1,1)$, on $\alpha$. Noting that $\alpha$ only influences $Z$ through the number of non-zero columns $K_i^+$, we compute

$$\begin{aligned}
p(\alpha|Z) &\propto p(Z|\alpha)p(\alpha) \\
&\sim \text{Poisson}\left(K^+; \alpha\left(\psi\left(1+\sum t\right)-\psi(1)\right)\right) \cdot \mathcal{G}(1,1) \\
&\sim \mathcal{G}\left(K^+ + 1, \psi\left(1+\sum t\right) - \psi(1) + 1\right),
\end{aligned}$$

where $\sum t$ is the total edge length in the tree.

## 4 AN APPLICATION TO CHOICE

Choice models play important roles in both econometrics (McFadden, 2000) and cognitive psychology (Luce, 1959). They describe what happens when people are given two or more options and are asked to choose one of them. In this section, we will restrict our attention to choices between pairs of objects, though the methods presented here can be applied more generally.

Even in the simple case of binary decisions, people's choices are not deterministic. The Elimination By Aspects (EBA) model is a popular attempt to explain this variation (Tversky, 1972). EBA hypothesizes that choices are based on a weighted combination of the features of objects. Keeping our earlier notation, let $Z$ be a feature matrix where $z_{ik} = 1$ if the $i^{\text{th}}$ object has the $k^{\text{th}}$ feature and $z_{ik} = 0$ otherwise. For each of the features, there is a corresponding weight $w_k$. The higher the weight, the more influence that feature has. The EBA model defines the probability of choosing object $i$ over object $j$ as

$$p_{ij} = \frac{\sum_k w_k z_{ik}(1-z_{jk})}{\sum_k w_k z_{ik}(1-z_{jk}) + \sum_k w_k(1-z_{ik})z_{jk}}. \quad (9)$$

For comparison with previous results (Görür et al., 2006) we assume extra noise in people's choices, with $\tilde{p}_{ij} = (1-\epsilon)p_{ij} + 0.5\epsilon$.

If $X$ is the observed choice matrix where $x_{ij}$ contains how many times object $i$ was chosen over object $j$, then for any given $w$ and $Z$, the probability of $X$ is

$$P(X|Z,w) = \prod_{i=1}^{N}\prod_{i<j}\binom{x_{ij}+x_{ji}}{x_{ij}}\tilde{p}_{ij}^{x_{ij}}(1-\tilde{p}_{ij})^{x_{ji}}. \quad (10)$$

If the number of features is known, Wickelmaier and Schmid (2004) showed how to estimate the weight vector and feature matrix. In general, though, the number of features is not known. Therefore, Görür et al. (2006) applied the IBP to this model in order to simultaneously infer the number of features, the feature matrix, and the weights of these features, and obtained improved performance over previous models.

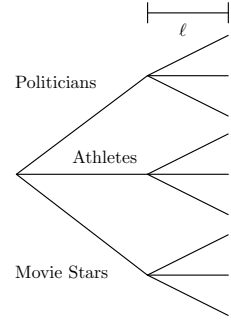

Figure 2: Hypothesis about a tree on preferences (Rumelhart and Greeno, 1971)

In an influential paper, Tversky and Sattath (1979) introduced the preference tree model as an extension of EBA. This model is applicable if the relationships of objects can be captured in a tree structure. In preference trees, each feature has to strictly obey the tree structure. That is, if two objects share a common feature, then all descendents of their most recent common ancestor must have that feature. In some situations, this tree structure may either be known in advance or a good working hypothesis may be available. An example can be found in an experiment reported by Rumelhart and Greeno (1971), in which subjects made 36 pairwise choices of who among a group of nine "well-known personalities" they would like to spend time with. The nine personalities consisted of three politicians, three athletes and three movie stars. It was therefore hypothesized that the tree structure summarizing the prior beliefs about these personalities was similar to the tree shown in Figure 2. In this figure, $\ell$ is the length of the edge from each general category of people to each individual at the leaf.

Just as the IBP can be used to infer features for EBA, the pIBP defines an appropriate prior for the case where features are assumed to follow a tree structure, as in preference trees. The pIBP model for feature generation can be seen as a soft version of preference trees, allowing features to break the tree structure but assigning low probability to these events. Görür et al. (2006) performed a comparison between the IBP as a prior and EBA models with fixed numbers of features as well as a finite preference tree model that was able to use the tree structure. It was shown that EBA with an IBP prior on the feature matrix outperformed all others. As we will show, using the pIBP gives both quantitatively and qualitatively better results in the case where the features are drawn using a tree.

To complete the full specification of the EBA model, we assume that each object has a unique feature as well as an unknown additional number of features that may or may not be shared as shown in Figure 3(a). We place a pIBP prior on the nonparametric part of the feature matrix and an independent $\mathcal{G}(1, 1)$ prior on each $w_k$. Inference proceeds as outlined above, with the addition of a Metropolis-Hastings step on $w_k$. A method similar to that in Görür et al. (2006) was used to deal with the $w_k$ values of new columns.

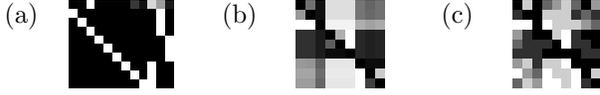

Figure 3: Example data demonstrating block structure. (a) True underlying features with corresponding weights in the top row. (b) Underlying probability choice matrix derived from (a) where the lighter the $(i, j)$ entry, the more likely $i$ is to be chosen over $j$. (c) An example observed choice matrix $X$ drawn from (b) with 5 observations per pair.

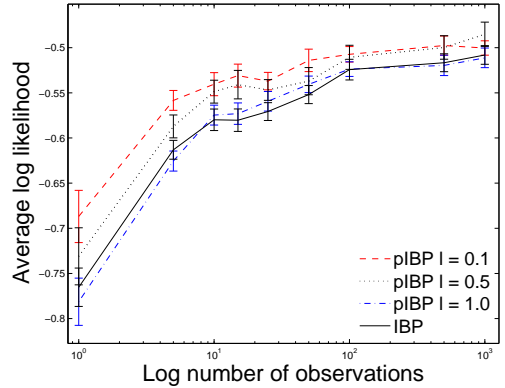

Figure 4: Comparison of the average predictive log-likelihood of the IBP and pIBP models with different degrees of prior knowledge along with error bars. On the x-axis, we vary the (log) number of observations seen for each $(i, j)$ pair.

We generated data from this choice model using the tree from Figure 2 with $\ell = 0.1$. The tree induces a "block structure" in the choice matrix, with the correlated features of objects along each branch resulting in similar patterns of choice for those objects. An example feature matrix generated from this model is shown in Figure 3(a). The top row shows the feature weights of the corresponding columns; the whiter the feature, the more weight that feature has. The feature matrix, $Z$, is displayed below where entries that are one are white and zero entries are black. Fifteen such examples were generated. For each of these examples, we computed the true value of choosing object $i$ over $j$ as shown in Figure 3(b) where the whiter the $(i, j)^{\text{th}}$ entry, the more likely $i$ is to be chosen over $j$. Based on these values, we generated data sets with 1, 5, 10, 15, 25, 50, 100, 500, and 1000 choice observations per pair $(i, j)$ following (9). An example of an observed matrix $X$ can be seen in Figure 3(c) with only 5 observations per pair. The lighter the $(i, j)^{\text{th}}$ entry, the more times $i$ was picked over $j$.

We used these data to examine the effects of two different factors. First, we wished to show the effect of using the pIBP prior over using the IBP prior as the number of observed choice decisions varied. The use of prior knowledge should always help, but with more observations, the influence of the prior should decrease. Second, we wished to test the effect of varying $\ell$ in our prior. The three values we tested were $\ell = 0.1$, $\ell = 0.5$, and $\ell = 1.0$. As mentioned in Section 2.3, the pIBP with $\ell = 1.0$ is the same as the IBP, but does not integrate over $\pi_k$ analytically.

For each of the nine observation levels on each of the fifteen examples, we performed leave-one-out cross validation with each model. For each model and validation point at each observation level, we ran an MCMC sampler for 3000 iterations from three different random initialization points. The first 1000 samples from each run were discarded as a burn-in period even though all chains appeared to have mixed within 100 iterations. The predictive likelihood was then averaged across every $10^{\text{th}}$ sample for all configurations. These results can be seen in Figure 4.

As expected, since the pIBP with $\ell = 0.1$ was using the true tree that generated the data, it was able to beat all other configurations for all numbers of observations except 1000, in which case all algorithms performed similarly. As the number of observations increases, the effect of the prior decreases and the models perform more similarly. The pIBP with $\ell = 0.5$ performs better than the IBP, but not as well as the pIBP with the correct tree. This shows that even without perfect knowledge of the tree structure, by inserting some information into the prior, we are able to outperform algorithms that cannot use the same prior knowledge. We also see that the IBP and pIBP with $\ell = 1.0$ perform nearly identically, so explicitly sampling $\pi_k$ in the inference algorithm does not influence the results.

In addition to obtaining higher likelihoods, the results from the pIBP were also more concise. In Figure 5(a), we show the average feature matrices for the pIBP and IBP when presented with the choice matrix in Figure 3(c). In order to obtain these averages, the $Z$ matrices for all samples after the burn-in period were collapsed into equivalent $Z$ matrices in which the weights of all identical columns were summed together. This results in the same probabilities under the EBA model and allows us to average these values across all samples.

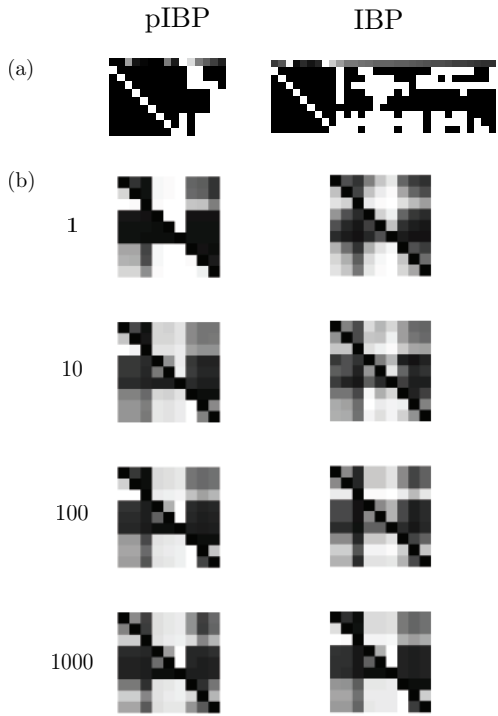

Figure 5: (a) Mean posterior $Z$ matrices with corresponding weights when presented with data from Figure 3. (b) Mean estimated choice matrices $P_{ij}$ for the pIBP and IBP when presented with the different numbers of observations from the data in Figure 3.

We also dropped all columns whose weight was below 0.1. As can be seen, the pIBP recovers a feature matrix very similar to the true data while the IBP requires many more features and still does not achieve the same performance. This large number of features necessary to explain the same data was also observed by Görür et al. (2006). Finally, we checked to see how many examples are needed for the choice probability matrix estimated from the samples of $Z$ and $W$ to show the same structure as the true choice probability matrix from Figure 3(b). In Figure 5(b), we show estimated choice matrices for 1, 10, 100, and 1000 observations per pair. With the pIBP, we observe a block structure immediately, though not all details of the choice matrix are correct. Within very few observations, though, the choice probabilities get close to the true values. In the IBP, we need more than 100 samples before it recovers the block structure.

## 5 CONCLUSION

We have described the Phylogenetic Indian Buffet Process, a novel non-exchangeable prior for infinite latent feature models. If we can summarize our prior knowledge about the relationships of objects using a tree structure, the pIBP allows us to perform nonparametric Bayesian inference efficiently. This allows us to incorporate our prior knowledge while still harnessing the power of nonparametric Bayesian methods to simultaneously infer both the number of features and their parameters. We have shown through an application to choice models that this can improve the performance of existing methods.

### Acknowledgements

This work was supported by the UC Berkeley Chancellor's Faculty Partnership Fund, by grant BCS-0631518 from the National Science Foundation and by a grant from the Lawrence Livermore National Laboratory.